\pdfoutput=1

\documentclass[11pt]{article}

\usepackage[]{EMNLP2024}

\usepackage{times}
\usepackage{latexsym}

\usepackage[T1]{fontenc}

\usepackage[utf8]{inputenc}

\usepackage{microtype}

\usepackage{inconsolata}
\usepackage{amsmath}
\usepackage{graphicx}
\usepackage{multirow}
\usepackage{latexsym}
\usepackage{amssymb}
\usepackage{makecell}
\usepackage{booktabs}
\usepackage{bbm}
\usepackage{wrapfig}
\usepackage{lipsum} 
\usepackage{xcolor}
\usepackage{caption}

\usepackage{color}
\usepackage{soul}
\usepackage[most]{tcolorbox}
\tcbuselibrary{skins}
\tcbuselibrary{breakable}
\usepackage{multicol}
\usepackage{adjustbox}
\usepackage{changepage}
\usepackage[fixed]{fontawesome5}

\usepackage{graphicx,calc}
\usepackage{wrapfig}
\newlength\myheight
\newlength\mydepth
\settototalheight\myheight{Xygp}
\usepackage{inconsolata}
\usepackage{CJKutf8}
\definecolor{my_green}{RGB}{51,102,0}
\definecolor{my_red}{RGB}{204, 0, 0}
\definecolor{my_purple}{RGB}{160, 43, 147}
\definecolor{my_blue}{RGB}{15, 158, 213}
\definecolor{my_yellow}{RGB}{245, 194, 66}

\newcommand{\ours}{\textsc{RuleAlign}}

\title{RuleAlign: Making Large Language Models Better Physicians  with Diagnostic Rule Alignment}

\author{Xiaohan Wang$^\spadesuit$$^\diamondsuit$\footnotemark[1], Xiaoyan Yang$^\heartsuit$\thanks{~~Equal contribution.}, Yuqi Zhu$^\spadesuit$$^\diamondsuit$, Yue Shen$^\heartsuit$, Jian Wang$^\heartsuit$\\
\textbf{Peng Wei$^\heartsuit$, Lei Liang$^\heartsuit$, Jinjie Gu$^\heartsuit$, Huajun Chen$^\spadesuit$$^\diamondsuit$, Ningyu Zhang$^\spadesuit$$^\diamondsuit$}\thanks{~~Corresponding author.}\\
$^\spadesuit$Zhejiang University,~ $^\heartsuit$Ant Group\\ 
$^\diamondsuit$ZJU-Ant Group Joint Research Center for Knowledge Graphs,\\
\texttt{\{wangxh07,zhangningyu\}@zju.edu.cn}\\
}

\begin{document}
\maketitle
\begin{abstract}
 
Large Language Models (LLMs) like GPT-4, MedPaLM-2, and Med-Gemini achieve performance competitively with human experts across various medical benchmarks. However, they still face challenges in making professional diagnoses akin to physicians, particularly in efficiently gathering patient information and reasoning the final diagnosis. To this end, we introduce the RuleAlign framework, designed to align LLMs with specific diagnostic rules. We develop a medical dialogue dataset comprising rule-based communications between patients and physicians and design an alignment learning approach through preference learning. Experimental results demonstrate the effectiveness of the proposed approach. We hope that our work can serve as an inspiration for exploring the potential of LLMs as AI physicians.

\end{abstract}

\def\ours{UrologyRD}

\section{Introduction}

Medical diagnosis involves physicians obtaining patients' subjective symptoms through questioning, as well as objective examination results, to infer the most likely diseases or health issues \cite{zhang2022cblue,he2023survey,gondocs2024ai,caruccio2024can}.
Large Language Models (LLMs) can extend medical resources by leveraging medical corpora in pretraining to simulate physicians, providing initial screening services that alleviate the burden on physicians and accelerate the diagnostic process for patients ~\cite{singhal2023large,DBLP:journals/npjdm/PengYCSPCMFZMLMOAHSGBW23,doi:10.1056/AIra2400012,zhang2023huatuogpt,thirunavukarasu2023large,bao2023disc}.
However, LLMs face a significant gap compared to real physicians~\cite{DBLP:journals/corr/abs-2311-05112,jiang2023health,DBLP:journals/corr/abs-2401-05654}, particularly in specialized inquiries for specialty diseases.

\begin{figure}
    \centering
    \includegraphics[width=0.95\linewidth]{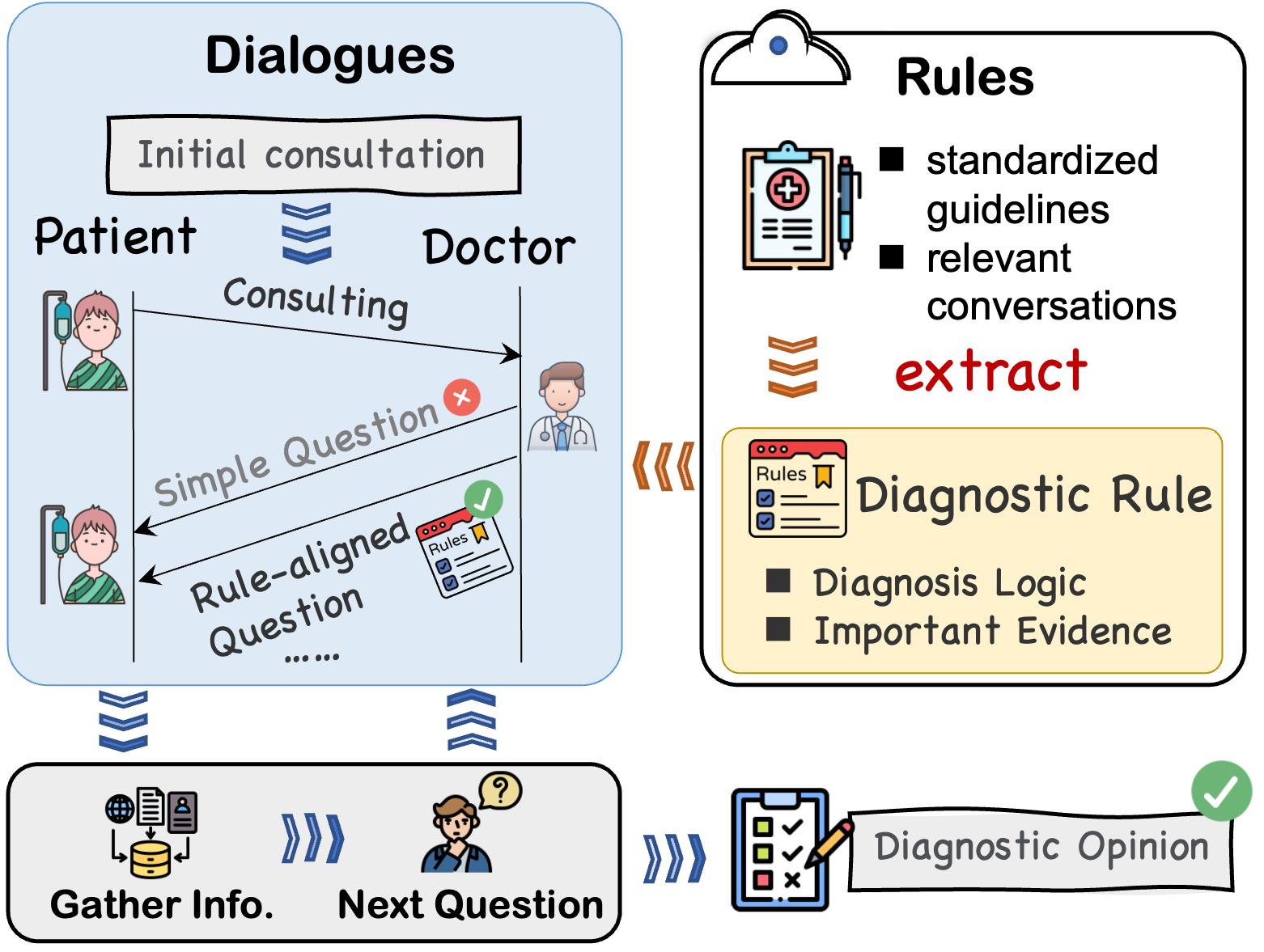}
    \caption{In medical practice, physicians need to gather sufficient patient information through inquiry to make the final diagnostic opinion.
    As professional physicians, their questions are usually rule-aligned, making the entire process efficient and logical.}
    \label{intro}
\end{figure}

The paradigms of existing works are too coarse in disease categories and differ greatly from actual consultations~\cite{DBLP:journals/corr/abs-2310-15896,DBLP:journals/bioinformatics/JinKCCYWL23,gao2023designing}.
One critical issue is logical consistency: LLMs may propose diagnostic hypotheses without adequate informational support or disregard previous replies, leading to unsatisfactory responses~\cite{DBLP:journals/patterns/LievinHMW24}.
Moreover, LLMs face significant challenges in adhering to rules~\cite{mesko2023imperative,DBLP:journals/corr/abs-2402-17358,luo2024asgea}. 
Note that medical practice is governed by stringent professional guidelines and standards as shown in Figure \ref{intro}, yet LLMs often demonstrate limited familiarity with regulations, resulting in informational shortfalls. 
Furthermore, the lack of specialized knowledge is evident as LLMs struggle to accurately comprehend and utilize professional terminology, 
rendering them incapable of providing convincing diagnoses and actionable advice.

To this end, we introduce the \textbf{RuleAlign} framework to make LLMs better physicians with diagnostic rule alignment. 
Specifically, we collect the necessary diagnostic rules within the field of urology and provide a rule-based dialogue dataset~\ours. 
It is designed to guide the behavior of LLMs, ensuring that their responses align with established protocols.
We train the model by constructing optimized preference pairs without additional human-annotation resource, which can achieve improvements in both single-round evaluation and multi-round Standard Patient testing~\cite{DBLP:journals/corr/abs-2403-16446}.

\begin{figure*}
    \centering
    \includegraphics[width=1\linewidth]{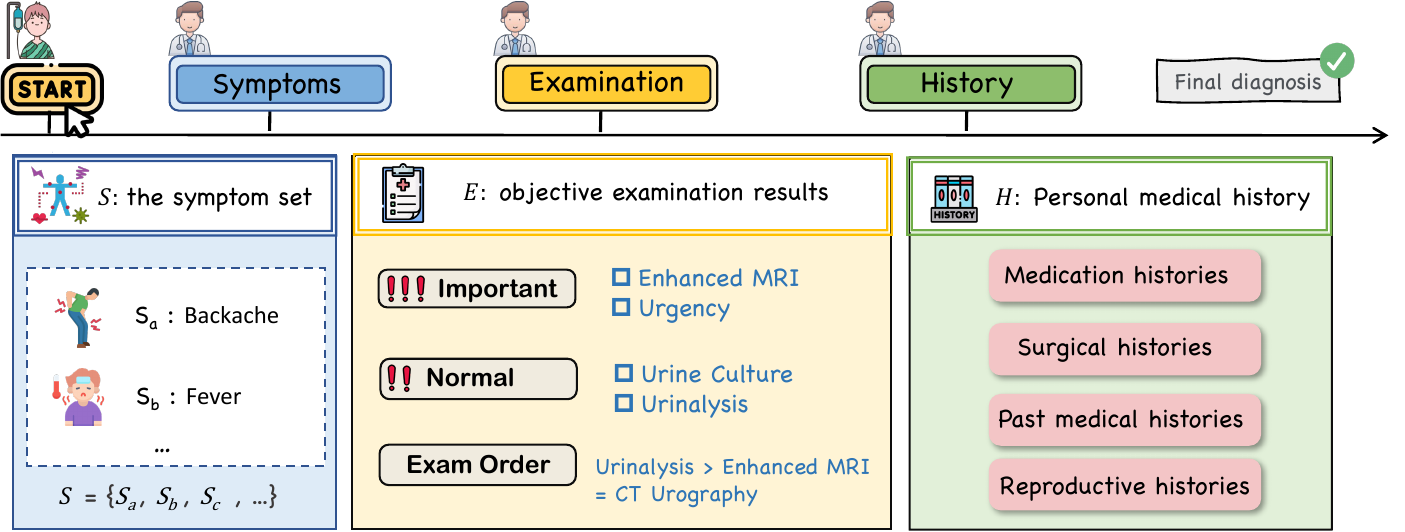}
    \caption{Physicians' diagnostic rules are specified for specific diseases and closely associated evidence in this study. 
    They adhere to not only general diagnostic principles but also specialized disease requirements.}
    \label{fig:enter-label}
\end{figure*}

\section{Related Works}
\paragraph{Medical LLMs.}
Medical LLMs hold significant application value, with both academic and industrial sectors actively advancing their development.

By introducing medical data into general LLMs, supervised finetuning (SFT) becomes more common, such as MedPaLM-2~\cite{DBLP:journals/corr/abs-2305-09617}, and Med-Gemini~\cite{ DBLP:journals/corr/abs-2404-18416}. 
It also leads to the emergence of many Chinese medical LLMs, including DoctorGLM~\cite{DBLP:journals/corr/abs-2304-01097}, HuatuoGPT-II~\cite{DBLP:journals/corr/abs-2311-09774}, and Zhongjing~\cite{DBLP:conf/aaai/YangZZZXJZ24}.
BianQue~\cite{DBLP:journals/corr/abs-2310-15896} utilize BianQueCorpus dialogue dataset optimized and constructed via ChatGPT, balancing the capability of asking questions and providing health advice. 
The characteristic of DISC-MedLLM~\cite{DBLP:journals/corr/abs-2308-14346} lies in enhancing conversational ability by incorporating knowledge graph question-answer pairs and human preference guided conversations into instruction data.
PLPF~\cite{DBLP:journals/corr/abs-2401-05695} involves rule modeling and preference learning to integrate the doctor’s diagnostic logic into LLMs.

\paragraph{Model Alignment.}

Human feedback~\cite{DBLP:conf/nips/Ouyang0JAWMZASR22} is widely applied to enhance the performance of LLMs through optimized preference learning.
RLHF~\cite{DBLP:conf/nips/ChristianoLBMLA17} employs a reward model trained via the Bradley-Terry (BT) model to suggest optimized outputs that maximize rewards through RL algorithms like PPO~\cite{schulman2017proximal}.
SLiC~\cite{DBLP:journals/corr/abs-2305-10425} ranks model-generated preferences by combining losses, while RRHF~\cite{DBLP:conf/nips/YuanYTWHH23} assumes multiple ranked responses for zero-margin likelihood contrastive loss; both methods rely on explicit reward models.
The propose of DPO~\cite{DBLP:conf/nips/RafailovSMMEF23} provides a theoretical and technical basis for alignment methods free from the dependence on reward models.
DPO directly uses the BT model to fit the preferences.
RSO~\cite{DBLP:journals/corr/abs-2309-06657} combines the benefit of SLiC and DPO with data augmentation through statistical rejection sampling.
SPIN~\cite{DBLP:journals/corr/abs-2401-01335} enhances DPO using self-iterative training with SFT data, treating synthetic data as rejection responses and high-quality SFT data as selection responses, iterating until the model can no longer distinguish between them.

\section{Dataset Construction}

\label{subsect:datadetail}
\subsection{Diagnostic Rule Collection}
To meet the needs of real patient clinical scenarios, we introduce more detailed diagnostic rules during the dataset construction. 
By summarizing relevant conversations and extracting key rules from standardized diagnostic guidelines, we specify the physician's diagnostic rules at the level of target diseases and strongly associated evidence as shown in Figure~\ref{fig:enter-label}. 
To further focus our research and better address specialized issues, we take urology as an example and construct rules for its common target diseases, that is, each \textbf{disease} $d_\textit{i}$ is equipped with its corresponding diagnostic \textbf{rules} $r_\textit{i}$.

Diagnostic rules $r_\textit{i}$ can be divided into two aspects: \textbf{constraints within the diagnosis logic} and \textbf{the search for important evidence}. 
The former at the logical layer primarily proceed according to the \textbf{trajectory} reformulated as $\tau_\textit{i} =(s_\textit{i} \rightarrow e_\textit{i} \rightarrow h_\textit{i} \rightarrow d_\textit{i})$.
Here, the trajectory components are defined as follows:  the patient's subjective symptom description $s_\textit{i}$, the collection of patient's objective examination results $e_\textit{i}$ , the inquiry into the patient's personal medical history $h_\textit{i}$, and the final diagnostic opinion. 
Patients often first report pain or discomfort; strategically following this trajectory can effectively collect more comprehensive medical information.

As for key patient evidence, patients frequently fail to identify which information is relevant for an accurate diagnosis, necessitating detailed inquiries by doctors.
Within the symptom set \(S = \{S_a, S_b, S_c, \ldots\}\), based on diagnostic knowledge, symptoms \(S_a\) and \(S_b\) are often associated with the target disease, making them key diagnostic indicators. For example, fever and difficulty urinating are important symptom evidences for bladder cancer. 
Objective examination results \(E = \{E_a, E_b, E_c, \ldots\}\), comprising physical exams, laboratory tests, and imaging studies, are essential for their reliability and minimal bias, thereby helps confirm the validity of a diagnosis. 
Professional doctors select and categorize examination evidence based on importance and provide a corresponding order $ER$ based on practical experience.
Personal medical history \(H\) can be categorized into medication, surgical, past medical, and reproductive histories; selective questioning based on necessity.
Therefore, \(K_\textit{i} = \{S_{d_\textit{i}}, E_{d_\textit{i}}, ER_{d_\textit{i}}, H_{d_\textit{i}}\}\) denotes the \textbf{essential patient evidences}, including symptoms, examination results, examination order and medical history.
In conclusion, the diagnostic rules $r_\textit{i}$ integrate the elements $\tau_\textit{i}$ and $K_\textit{i}$ to enhance the efficiency and effectiveness of the diagnostic process. 

\subsection{Dataset Generation}
\label{sec:datageneration}

\paragraph{Data Collection.}
Open-sourced medical dialogue datasets~\cite{zeng-etal-2020-meddialog} frequently lack structured examination results, leading to incomplete diagnostic trajectory. 
Due to the overlap between disease categories and our specific targets, we choose the RJUA-QA~\cite{DBLP:journals/corr/abs-2312-09785} as initial data. 
After filtering the categories, we convert single-turn question $q$ and corresponding disease $d$ into dialogues $M$, encompassing the trajectory from initial symptoms to physician diagnoses, represented as \(M = \textit{G}_M(q , d)\), where $\textit{G}$ denotes the GPT-4 turbo API.

\paragraph{Disease Name Mapping.}
In data construction, it is imperative to map disease names meticulously. 
This involves distilling intricate pathology descriptions found in clinical records into broader medical categories (e.g., "right renal hydronephrosis" to "renal hydronephrosis"). 
Details like which side and progression stage while critical for clinical investigations, wield limited influence on query and risk complicating study designs unduly. 
Consequently, we map disease terms to defined categories, ensuring robust data compilation and precise alignment with established diagnostic rules.

\paragraph{Diagnostic Rule Adaption.}
By applying rules $r$ into the prompts for LLMs, we transform the original dialogue $M$ into a rule-based dialogue $M^r$. This adaptation process can be  formulated  as:
\[M^r_\textit{i} = \textit{G}_p(M_\textit{i}, r_\textit{i}) = \textit{G}_p(\textit{G}_M(q_\textit{i}, d_\textit{i}), \textit{T}(\tau_\textit{i},K_\textit{i}))\]
where $\textit{T}$ represents the template of rules and  $p$ encapsulates the constraints necessary to ensure the patient's honesty and clarity of intent. 
Through this approach, the synthesized dialogues not only conform more closely to the prescribed diagnostic rules but also maintain consistency and reliability.

\subsection{Quality Control}

To ensure precision and integrity, \ours\ undergoes rigorous annotation by professional doctors. 
In our process, urology specialists review all diagnostic rules $r_\textit{i}$. 
Based on these validated rules, the pass rate for a hundred samples  in 
 \ours\ approaches 70\%, with experts deeming these dialogues to be comprehensive and logical.
 We list the data statistics in Table \ref{datastatis}.

\begin{table}[ht]
\centering
\resizebox{1.0 \columnwidth}{!}{
\begin{tabular}{c l c}
\hline
Statistics& Items & Number\\
\hline
\multirow{6}{*}{Number}  & \textit{Disease} & 32\\
        & \textit{Diagnostic Rules} & 32\\
        & \textit{Train Dialogues} &  2,267\\
        & \textit{Train Rounds}  &  14,599 \\
        & \textit{Test Dialogues} &  107\\
        & \textit{Test Rounds}  &  720 \\
\midrule
\multirow{2}{*}{Max / Min}   & \textit{Rounds Per Dialogue}&  13/ 3\\
        & \textit{Length Each Round}&   200/3 \\
\hline
\end{tabular}
}
\caption{This table outlines the statistics for \ours, including counts and lengths for various components.}
\label{datastatis}
\end{table}
\begin{figure}
    \centering
    \includegraphics[width=1\linewidth]{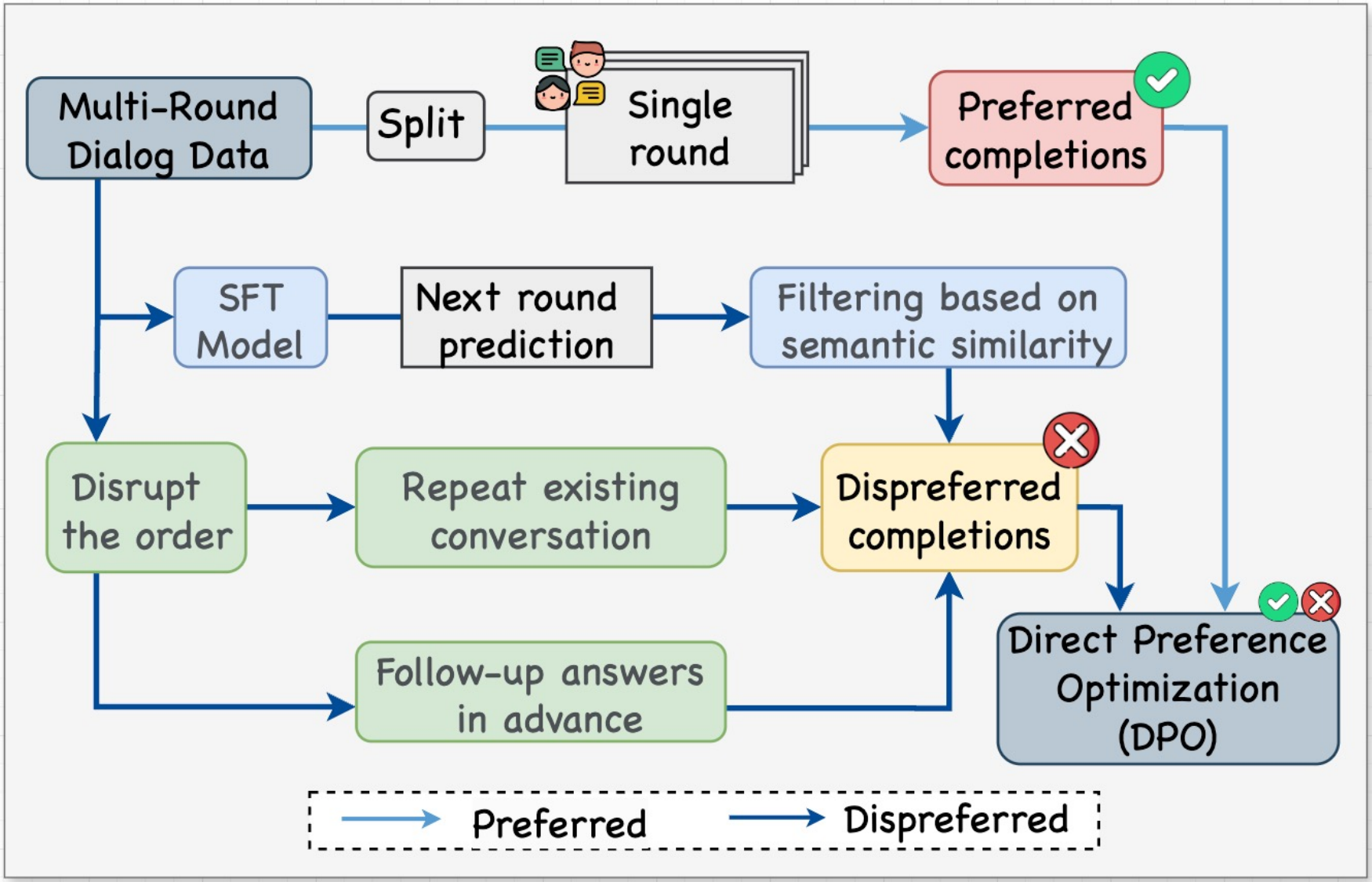}
    \caption{The pipeline of \textbf{RuleAlign}. The optimization contains distinct strategies to build the preference pairs without extra human-annotation resource.}
    \label{fig:rule_gragh}
\end{figure}

\section{The Proposed Approach: RuleAlign}

\subsection{Overview}
Preference learning~\cite{jiang2024survey}, which leverages human-labeled preference pairs, is considered an effective method for aligning LLMs with human objectives. 
DPO stands as a prominent approach in offline preference optimization~\cite{DBLP:conf/nips/RafailovSMMEF23}, notable for eliminating the need for an explicit reward model. 
High-quality preference pairs are essential for DPO but typically demand substantial manual annotation resources.

\subsection{Preference Learning}
The \ours\ dataset $\mathcal{D}$ encompasses both precise inquiry information and complex diagnostic rules, forming a comprehensive resource for disease diagnosis. 
The progression from pre-trained models to aligned LLMs using the DPO method generally involves two distinct phases: training the policy model through supervised fine-tuning, and conducting preference optimization.

In SFT phase, the language model is finetuned with high-quality instruction data $\mathcal{D}$ closely aligned with the objectives of the downstream task.
This includes instructions from users' inputs $x$ accompanied by suitable responses $y$.
The LLM $\pi_\theta$ generates a probability distribution $r_\theta(x,y)$, optimized by minimizing the negative log-likelihood.

\begin{equation}\label{loss:sft}
  \begin{aligned}
  \mathcal{L}_{\text{SFT}}(\pi_\theta)=&
  -\underset{\langle x,y\rangle\sim \mathcal{D}}{\mathbb{E}}
    \log \sigma(
    r_\theta(x,y)
)
  \end{aligned}
\end{equation}

In the second phase, the SFT model serves as the reference policy $\pi_{\text{ref}}$, while the policy model $\pi_\theta$ acts as the trainable parameters for preference learning. 
Unlike the prior RLHF methods, DPO calculates the optimization strategy's reward using a closed-form expression, implicitly defined by the models $\pi_{\theta}$ and $\pi_{\text{ref}}$, denoted as

\begin{equation}
r_\theta(x,y)=\beta\log\frac{\pi_\theta(y|x)}{\pi_\text{ref}(y|x)}
\end{equation}

Here, $\beta$ signifies a hyper-parameter controlling the scale of the reward difference.
DPO directly simplifies the learning of the optimal strategy using preference pairs $\mathcal{D}^p$, aiming to increase the likelihood of preferred completions $y_w$ and decrease the likelihood of dispreferred completions $y_l$, as
\begin{equation}\label{loss:dpo}
  \begin{aligned}
  \mathcal{L}_{\text{DPO}}(\pi_\theta;\pi_{\text{ref}})=&
  -\underset{\langle x,y_w,y_l\rangle\sim \mathcal{D}^p}{\mathbb{E}}\bigl[\\
    &\log \sigma(
    r_\theta(x,y_w) - r_\theta(x,y_l)
)  \bigr]
  \end{aligned}
\end{equation}

\subsection{Preference Pair Optimization}
We propose an efficient method, \textbf{RuleAlign}, which is  designed to automatically generate and optimize preference data for alignment (in Figure~\ref{fig:rule_gragh}). 
Building on previous research, we utilize dialogues synthesized using diagnostic rules to represent positive preferences, while responses after the SFT stage are designated as dispreferred for DPO. 
When the LLM output $y_l$ closely matches $y_w$, the loss approaches zero. 
Conversely, a greater disparity results in increased loss and larger gradient updates, thereby enhancing the learning of preferences.

However, the experiment reveals that original preference pairs do not consistently improve the performance of LLMs. 
In fact, models refined using DPO can occasionally perform worse than those solely optimized through SFT.
This exploration underscores the critical need to match the dispreferred completions with these less satisfactory instances for efficient learning.
Consequently, we propose an approach to enhance dispreferred completions $y_l$ through sample filtering using semantic similarity and the deliberate disruption of dialogue order.
Specifically, the former entails multiple sampling outputs after SFT and selecting the item with the lowest similarity to $y_w$ by BLEU score and lower than the threshold.
Additionally, the dispreferred completions are constructed by repeating existing responses or using the subsequent response to disrupt the diagnostic logic.

\section{Experiments}
\begin{table*}[ht]
\centering
\resizebox{2.0\columnwidth}{!}{
\begin{tabular}{llcccc}
\toprule
Model & Method & Perplexity$\downarrow$ & ROUGE-1 / 2 / L$\uparrow$ & BLEU$\uparrow$ & Length Rate (1) \\
\midrule
 & Base  & 15.78 & 21.61 / 4.45 / 16.79 & 4.86 & 4.29 \\
Baichuan2-7B-chat & SFT & 3.69 & 43.91 / 18.81 / 38.80 & \textbf{20.07} & 1.11 \\
 & DPO & \textbf{3.68} & 37.75 / 17.96 / 37.75 & 18.78 & 1.10 \\
 & RuleAlign & 3.77 & \textbf{44.44} / \textbf{19.04} / \textbf{39.30} & 19.87 & \textbf{1.07} \\
\midrule
 & Base  & 49.44 & 13.74 / 2.01 / 8.54 & 2.24 & 9.04 \\
Qwen1.5-7B-chat & SFT & 4.37 & 33.14 / 12.23 / 25.27 & 11.43 & 5.15 \\
 & DPO & \textbf{3.19} & 42.98 / 18.05 / 38.00 & 18.14 & 0.97 \\
 & RuleAlign & 3.28 & \textbf{44.20} / \textbf{18.82} / \textbf{39.21} & \textbf{19.48} & \textbf{1.03} \\
 \midrule
 & Base  & 20.92 & 15.17 / 2.69 / 25.26 & 1.96 & 14.49 \\
Huatuo-II-7B & SFT & 3.74 & 44.18 / 19.34 / 39.19 & 20.18 & 1.10 \\
 & DPO & \textbf{3.72} & 42.49 / 17.19 / 37.53 & 18.33 & 1.11 \\
 & RuleAlign & 3.81 & \textbf{44.53} / \textbf{19.45} / \textbf{39.67} & \textbf{20.26} & \textbf{1.06} \\
\bottomrule
\end{tabular}
}
\caption{We compare the performance of different models in single-round test. The best results are marked in \textbf{bold}.}
\label{tab:comparison}
\end{table*}

\subsection{Settings}
\paragraph{Evaluation Scenarios.}
When examining LLMs as physicians, the standardization of patient responses presents a challenge due to the inherently history-dependent nature. 
Therefore, it is recommended that model evaluations be conducted separately in both single-round and multi-round settings to more effectively assess their competencies.
In single-round test, the test set is selected from previously constructed dataset, using target dialogue rounds as input/output pairs with prior rounds as history.
In multi-round context, we find that Standardized Patient (SP) Testing is a requisite and commonly employed methods for assessing clinical dialogue. 
We follow the framework proposed by~\citet{DBLP:journals/corr/abs-2403-16446} covering diseases we involve.

\paragraph{Baselines.}
Three Chinese LLMs are chosen as base models. 
Baichuan2-7B-chat~\cite{baichuan2023baichuan2} and Qwen1.5-7B-chat~\cite{qwen} represent widely accessible open-source LLMs, exemplifying distinct architectures.
While Huatuo-II-7B~\cite{DBLP:journals/corr/abs-2311-09774} as a  medical LLM is predicated upon the foundations of Baichuan2-7B-Base, augmented with a corpus of medical SFT data.

\paragraph{Metrics.}
In single-round test, several metrics - \textbf{perplexity, Rouge and BLEU series, and Length Rate} - serve as key evaluative tools.
Perplexity assesses the likelihood of LLMs generating a specific sequence, with lower values suggesting better training. 
ROUGE and BLEU metrics evaluate the semantic and syntactic similarity of generated text with ground-truth, where higher values reflect greater congruence. 
Lastly, Length Rate measures the proportional length of generated text to ground-truth, near 1.0 indicating closer alignment.

In SP testing, we adopt metrics in the framework, encompassing five critical dimensions of diagnosis: \textbf{Information Completeness}, \textbf{Guidance Rationality}, \textbf{Diagnostic Logicality}, \textbf{Clinical Applicability} and \textbf{Treatment Logicality}. 
Collectively, these metrics assess the effectiveness of LLMs from the initial gathering of patient information to the logical reasoning behind diagnostic opinions and the practical execution of reaching a conclusive diagnosis, ensuring a comprehensive evaluation in simulating reasoning and decision-making processes.

\subsection{Main Results}

We report the main results of the experiments in Table \ref{tab:comparison}, highlighting the superior performance of RuleAlign.
Applying RuleAlign to three distinct models demonstrates its effectiveness in aligning LLMs with medical diagnosis task, resulting in more accurate and natural responses. 
RuleAlign reduces the perplexity to a range of 3 to 4, while also improving the ROUGE-1 by 20-30 and increasing the BLEU to nearly 20. 
By leveraging the instructions, LLMs are capable of fitting rules; however, there are still certain constraints in the model post-SFT.
On the one hand, there exists hallucinations after SFT, and in the case of Qwen, it continues to output irrelevant content after necessary outputs, resulting in longer responses and lower scores.
Employing DPO with direct SFT outputs and targets can alleviate such hallucinations.
Thus, the performance of DPO is superior to that of SFT in Qwen.
On the other hand, DPO may underperform SFT, since the disparities have the potential to confuse the LLMs in alignment. 
When it comes to Baichuan and Huatuo, there are cases of poor performance in terms of specific logic following where the order of inquiry is either repeated or skipped. 
It is noted that RuleAlign, optimizing dispreferred completions by semantic similarity filtration and order disruption, could improve the quality of generated text in metrics such as Rouge and BLEU, as well as Length Rate, despite a minor shortfall in perplexity.
What's more, the comparison between Huatuo and Baichuan demonstrates that normal medical corpus show little affect on diagnoses or further alignment on diagnostic rules.

\subsection{Analysis}

\begin{figure}
    \centering
    \includegraphics[width=0.98\linewidth]{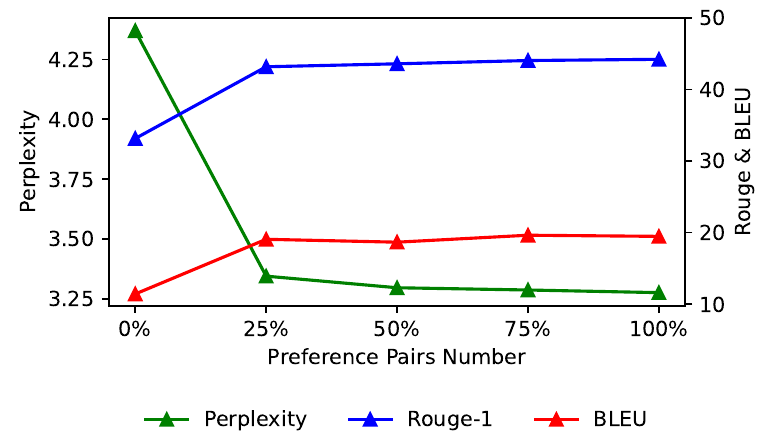}
    \caption{The results for preference pairs of different number sizes performed in Qwen1.5-7B-chat.}
    \label{fig:ablation-num}
\end{figure}

As displayed in Table \ref{tab:action_analysis} and Figure \ref{fig:ablation-num}, we conduct further analysis on the primary components of preference pairs that affect on model's ability to predict the next round of inquiry.
To verify the validity of strategies in preference pair optimization, we compare RuleAlign with only applying semantic similarity filtration or order disruption to build the dispreferred completions. 
The ablation study reveals that adding both strategies of RuleAlign could bring about better performance in metrics Rouge and BLEU.
Also, it explains that the unsatisfied rise of the perplexity score in RuleAlign is attributed to the introduction of order disruption.
In addition, the scale of preference pairs, depicted in Figure \ref{fig:ablation-num}, shows that a quarter of the preference pairs are sufficient to train the SFT model and learn about the preference.
Increasing the number of preference pairs on this basis does not bring significant improvement to the effect.
Thus, for the rule alignment of medical diagnosis, the strategy of designing preference pairs has a more important impact on the accuracy of single-round prediction.

\begin{table}[tb]
\centering
\resizebox{1\columnwidth}{!}{
\begin{tabular}{lccc}
\toprule
Qwen1.5 & Perplexity & Rouge-1 & BLEU \\
\midrule
all preference pairs & 3.28 & 44.20 & 19.48 \\
- only similarity & 3.19 & 42.91 & 18.66 \\
- only disruption &  3.53 &  43.09 &  18.97\\
\bottomrule
\end{tabular}}
\caption{The result of different preference pair optimization strategies performed in Qwen1.5-7B-chat.}
\label{tab:action_analysis}
\end{table}

\subsection{Standardized Patient (SP) Testing}

This sophisticated multi-agent system is explicitly designed to emulate the dynamic dialogues between patients and the physician agent. 
This incorporates Retrieval-Augmented Evaluation (RAE) mechanism that predicts patient responses based on the repository that archives a vast array of symptoms and test outcomes, thereby providing a realistic simulation of patient behavior.

\begin{figure}
    \centering
    \includegraphics[width=1.0\linewidth]{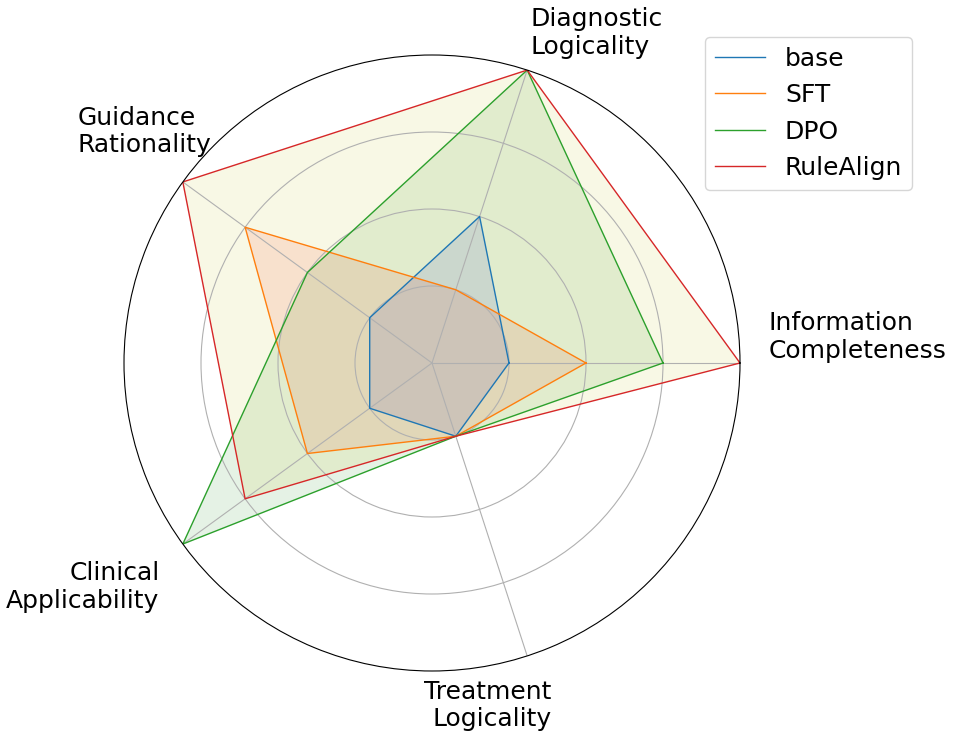}
    \caption{This radar plot shows the SP testing ranks of different methods, with closer to the outer edge indicating higher ranking and better performance.}
    \label{fig:sp-result}
\end{figure}
\paragraph{SP Results.}
Figure \ref{fig:sp-result} illustrates RuleAlign's multidimensional abilities to communicate with patients that mimic real-world diagnoses and rank the scores among different methods.
The increase of information completeness mirrors the advancements of LLMs in collecting sufficient critical patient information during conversations.
Especially, utilizing evidence knowledge in diagnostic rules, RuleAlign instructs LLMs to query about more targeted test and examination reports which brings about the improvement of guidance rationality.
Additionally, the rise in diagnostic logicality measures how the model logically deduces accurate diagnostic conclusions in accordance with the accessible information.
RuleAlign also reaches over 5 turns dialogues in average in metric clinical applicability.
But in the absence of treatment knowledge enhancement, all four approaches are tied in terms of metric treatment logicality. 
Meanwhile, RuleAlign still exists a great disparity in delivering final accurate diagnoses and treatment suggestions comparable to those provided by real-world physicians.
We illustrate a case of SP testing in Figure~\ref{fig:case}  to demonstrate the effectiveness of RuleAlign.

\section{Conclusion}
In this study, we build the medical dialogue dataset \ours\ based on the diagnostic rules and propose the innovative method RuleAlign to automatically synthesize the preference pairs for alignment. 
The experiment results show the effectiveness of RuleAlign on various evaluation settings.
And we anticipate our work will bring benefits to further research on LLMs as medical applications or AI physicians.

\section*{Ethical Considerations}

This work adheres strictly to ethical standards and best practices in research.
All medical data utilized are extracted from publicly available datasets that do not contain any sensitive or proprietary patient information. 
Consequently, this research is conducted without any ethical concerns.

\section*{Acknowledgements}
We would like to express our sincere gratitude to LLaMA-Factory~\cite{zheng2024llamafactory} for efficient fine-tuning pipelines.
Our appreciation also goes to RJUA-QA~\cite{DBLP:journals/corr/abs-2312-09785} and RJUA-SP~\cite{DBLP:journals/corr/abs-2403-16446}.

\bibliography{anthology,custom}
\bibliographystyle{acl_natbib}
\appendix
\clearpage

\section{Appendix}
\label{sec:appendix}

\subsection{Training details}
In our experiment, we employ an NVIDIA A100 GPU to train the aligned models.
During both the SFT stage and the preference alignment phase, we incorporate the LoRA technique in the training process. 
The detailed hyperparameters we use are shown in Table~\ref{tab:hyperparameters} and Table~\ref{tab:dpohyperparameters}.

\begin{table}[!htp]
    \centering
    \renewcommand\arraystretch{1.1}
    \caption{Detailed hyperparameters used in SFT}
    \begin{tabular}{c|c}
        \toprule
        \textbf{Name} & \textbf{Value} \\
        \midrule
        lora rank & 8 \\
        lora alpha & 16 \\
        lora target & all \\
        cutoff len & 1,024 \\
        epochs & 10 \\
        batch size per device & 4 \\
        gradient accumulation steps & 8 \\
        learning rate & 5e-5 \\
        \bottomrule
    \end{tabular}
    \label{tab:hyperparameters}
\end{table}

\begin{table}[!htp]
    \centering
    \renewcommand\arraystretch{1.1}
    \caption{Detailed hyperparameters used in RuleAlign and DPO.}
    \begin{tabular}{c|c}
        \toprule
        \textbf{Name} & \textbf{Value} \\
        \midrule
        lora rank & 8 \\
        lora alpha & 16 \\
        lora target & all \\
        cutoff len & 1,024 \\
        epochs & 1 \\
        batch size per device & 4 \\
        gradient accumulation steps & 8 \\
        learning rate & 2e-5 \\
        \bottomrule
    \end{tabular}
    \label{tab:dpohyperparameters}
\end{table}

\subsection{Dataset details}

\begin{figure}
    \centering
    \includegraphics[width=1.0\linewidth]{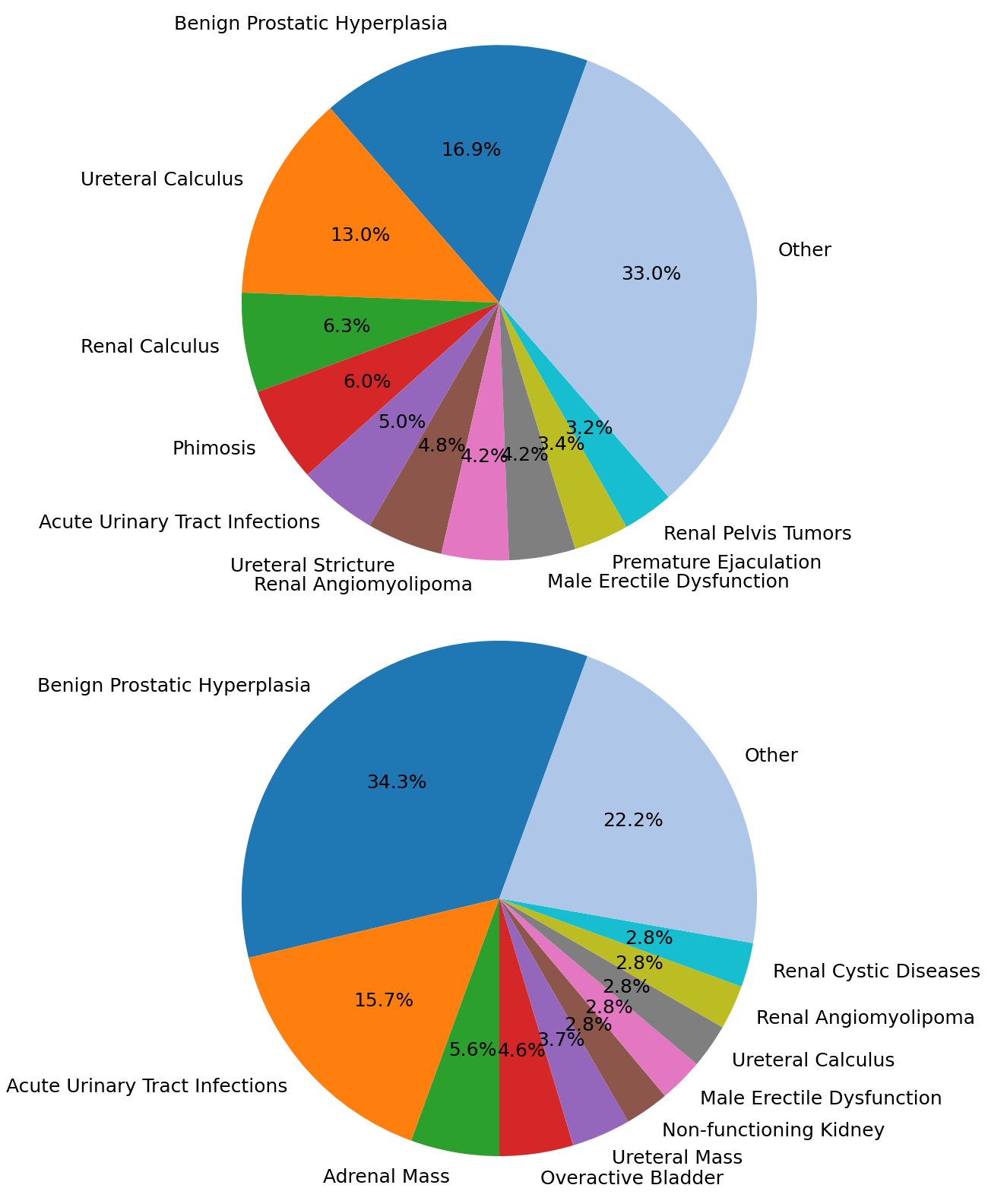}
    \caption{These plots illustrate the top 10 categories in train set (up) and test set (down) in \ours.}
    \label{fig:categ}
\end{figure}

This subsection expands §\ref{subsect:datadetail} with additional details about our data construction steps and statistical analysis.
The initial Q\&A dataset RJUA-QA covers 67 common urological disease categories,which is developed in collaboration with
department of urology Shanghai Renji Hospital~\cite{DBLP:journals/corr/abs-2312-09785}.
Through the construction pipeline, the \ours\ consists of train and test splits and spans across 32 diseases.
Figure~\ref{fig:categ} illustrate the disease categories distribution in \ours.
Figure~\ref{fig:datasetcase} shows an example of \ours\ dataset, a simulated diagnostic communications between patient and physician, which follows the designed rules for diseases.

In addition, the template prompt used as input for GPT-4 turbo API as introduced in section  \ref{sec:datageneration}, follows the formulas $G_M$ and $G_p$.
It should be mentioned that since the medical data used in this work are originally in Chinese, and we employ GPT4o for translation directly.

\subsection{SP Testing details}

\begin{table}[ht]
\centering

\label{tab:sp}
\resizebox{1.0\columnwidth}{!}{
\begin{tabular}{lcccc}
\toprule
SP Metric & Base & SFT & DPO & RuleAlign\\
\midrule
Information Completeness &6.04 & 15.14& 15.72 & 17.32 \\
Diagnostic Logicality&22.72 & 20.45 & 31.82& 31.82\\
Guidance Rationality&9.30  & 26.43& 25.36& 29.29 \\
Clinical Applicability&1.77  & 4.42 & 5.41 & 5.39\\
Treatment Logicality&4.55 & 4.55 & 4.55& 4.55\\

\bottomrule
\end{tabular}
}
\caption{SP comparison of  methods.}
\label{tab:spscores}
\end{table}

The SP testing framework offers necessary metrics to scrutinize whether the conduct exhibited by the physician agent adheres to the stringent protocols of professional clinical practice pathways.
Table~\ref{tab:spscores} presents the detailed metric scores for various methods employed within the SP testing framework. 
Given the variability in the range of each metric and the inclusion of 44 standard patients, the ranks of SP metrics in Figure \ref{fig:sp-result} are considered a more effective presentation for evaluation. 
\newpage
\begin{figure*}[!htp]
\centering
\scalebox{1.}{
\begin{tcolorbox}[center,breakable,title=Case Study in SP Testing]

\begin{multicols}{2}

\begin{center}
\large \textbf{English}
\end{center}
\textcolor{my_yellow}{\textbf{Patient}}: Doctor, I've had a dull pain on the right side of my waist for several months.\\
\textcolor{my_blue}{\textbf{Doctor}}: Hello, is this pain constant or intermittent? Is there anything that makes it worse?\\
\textcolor{my_yellow}{\textbf{Patient}}: No fever.\\
\textcolor{my_blue}{\textbf{Doctor}}: I see. Have you had any urinary system examinations, such as a routine urinalysis or a urinary system ultrasound?\\
\textcolor{my_yellow}{\textbf{Patient}}: One month ago, urinary system ultrasound showed: Strong echo of 15*12mm in the upper segment of the right ureter, with hydronephrosis.
Cardiac ultrasound, left atrial enlargement, left ventricular diastolic dysfunction, routine urinalysis one day before surgery: Urine color: straw yellow, urine clarity: clear, urine specific gravity: 1.028, urine pH: 5.5, urine nitrite: negative, urine leukocyte esterase: negative, urine protein: 2+↑, urine glucose: 4+↑, urine ketone bodies: 1+↑, urine urobilinogen: normal, urine bilirubin: negative, urine occult blood: 3+↑, microscopic examination ...\\
\textcolor{my_blue}{\textbf{Doctor}}: Based on the information and results you've provided, along with your symptoms, you might have hydronephrosis caused by a ureteral stone. We need further imaging studies to confirm the exact location and size of the stone. Have you had a more detailed imaging, such as a kidney and ureter CT scan?\\
\textcolor{my_yellow}{\textbf{Patient}}: I don't know, please ask other questions, doctor!\\
\textcolor{my_blue}{\textbf{Doctor}}: Understood. Do you have any other past medical history, such as hypertension, diabetes, or heart disease?\\
\textcolor{my_yellow}{\textbf{Patient}}: Hypertension, negative. Diabetes, positive.\\
\textcolor{my_blue}{\textbf{Doctor}}: Got it. Regarding diabetes, are you currently taking any medications to control your blood sugar?\\
\textcolor{my_yellow}{\textbf{Patient}}: No medication history related.\\
\textcolor{my_blue}{\textbf{Doctor}}: Okay, thanks. Based on the symptoms you described and the existing test results, combined with your history of diabetes, I believe you are very likely to have hydronephrosis caused by a ureteral stone on the right side. We need to conduct a kidney and ureter CT scan to determine the exact location and size of the stone for further diagnosis.

\columnbreak

\begin{center}
\large \textbf{Chinese}
\end{center}
\begin{CJK*}{UTF8}{gbsn} 
\textcolor{my_yellow}{患者}：医生您好，我右侧腰隐隐痛了好几个月。\\
\textcolor{my_blue}{医生}：你好，请问这个腰痛是持续性的还是间歇性的？有没有什么事情会让它加剧？\\
\textcolor{my_yellow}{患者}：无发热\\
\textcolor{my_blue}{医生}：了解了，你有没有做过泌尿系统方面的检查，比如尿常规或者泌尿系统B超？\\
\textcolor{my_yellow}{患者}：1月前泌尿系统B超检查：右侧输尿管上段强回声15*12mm，伴肾盂积水\\
心彩超，左房内径增大，左室驰张功能减退，术前1天尿常规，尿颜色 稻黄色，尿浑浊度 清亮，尿比重 1.028，尿酸碱度 5.5，尿亚硝酸盐 neg，尿白细胞酯酶 neg，尿蛋白质 2+↑，尿葡萄糖 4+↑，尿酮体 1+↑，尿胆原 norm，尿胆红素 neg，尿潜血 3+↑，镜检红细胞 100+/HP，镜检白细胞 未查见/HP，1天前\\
\textcolor{my_blue}{医生}：根据你提供的信息和检查结果，结合你的症状，你可能有输尿管结石引起的肾盂积水。我们还需要进一步的影像学检查来确认结石的确切位置和大小。你是否已经进行了更详细的影像学检查，例如肾输尿管CT？\\
\textcolor{my_yellow}{患者}：我不知道，请医生询问其他问题！\\
\textcolor{my_blue}{医生}：明白了。请问你有没有其他的既往病史，比如高血压、糖尿病或心脏病？\\
\textcolor{my_yellow}{患者}：高血压，阴性 糖尿病，阳性\\
\textcolor{my_blue}{医生}：了解了。关于糖尿病，你目前在服用什么药物控制血糖吗？\\
\textcolor{my_yellow}{患者}：没有相关用药史。\\
\textcolor{my_blue}{医生}：好的，感谢你提供的信息。根据你描述的症状和已有的检查结果，加上你的糖尿病病史，我认为你很可能患有右侧输尿管结石导致的肾盂积水。我们需要进行肾输尿管CT检查来确定结石的确切位置和大小，以便进一步确诊。\\
\end{CJK*}

\end{multicols}

\end{tcolorbox}

}
\caption{
This is a case of diagnostic consultation between our trained AI physician and the patient simulated by RJUA-SP framework. 
}
\label{fig:case}
\end{figure*}

\begin{figure*}[!htp]
\centering
\scalebox{1.}{
\begin{tcolorbox}[center,breakable,title=Example of Dataset \ours]
\begin{multicols}{2}

\begin{center}
\Large \textbf{English}
\end{center}
\textcolor{my_yellow}{\textbf{Patient}}: Doctor, hello. I am a 66-year-old male patient, and recently, during a check-up, I discovered a cyst on my right adrenal gland that has been there for 8 years.\\
\textcolor{my_blue}{\textbf{Doctor}}: Hello! Have you experienced any discomfort recently, such as weight changes, fatigue, or blood pressure fluctuations?\\
\textcolor{my_yellow}{\textbf{Patient}}: No special sensations, it was only found during a routine check-up.\\
\textcolor{my_blue}{\textbf{Doctor}}: Have you previously had an adrenal ultrasound examination?\\
\textcolor{my_yellow}{\textbf{Patient}}: No, I have not had an adrenal ultrasound.\\
\textcolor{my_blue}{\textbf{Doctor}}: Have you had an enhanced CT or MRI of the upper abdomen?\\
\textcolor{my_yellow}{\textbf{Patient}}: According to the CTA examination last month, I have multiple round cystic shadows on both kidneys, the largest on the right kidney being about 1 cm, and a round mass with a diameter of 3 cm on the right adrenal gland.\\
\textcolor{my_blue}{\textbf{Doctor}}: Okay, have you undergone any blood adrenal hormone tests or urine adrenal hormone tests?\\
\textcolor{my_yellow}{\textbf{Patient}}: No, I have not had those hormone tests.\\
\textcolor{my_blue}{\textbf{Doctor}}: Understood. Do you have a history of hypertension or any other past medical history?\\
\textcolor{my_yellow}{\textbf{Patient}}: I have a history of hypertension, but no diabetes, coronary heart disease, or cerebrovascular disease.\\
\textcolor{my_blue}{\textbf{Doctor}}: Got it. How is your hypertension controlled? Are you on medication?\\
\textcolor{my_yellow}{\textbf{Patient}}: I have been on medication to control it, and my blood pressure is quite stable.\\
\textcolor{my_blue}{\textbf{Doctor}}: Based on the information you have provided so far, along with your history of hypertension and the CTA examination results, we cannot rule out the possibility of an adrenal tumor. Although there are currently no clear symptoms of abnormal adrenal hormone secretion, to make a definitive diagnosis, we need to conduct blood adrenal hormone tests and urine adrenal hormone tests, as well as a possible enhanced MRI of the upper abdomen. These tests will help us determine the nature and functional activity of the tumor.

\columnbreak

\begin{center}
\Large \textbf{Chinese}
\end{center}
\begin{CJK*}{UTF8}{gbsn} 
\textcolor{my_yellow}{患者}：医生，您好。我是一名66岁的男性患者，最近我做检查发现右肾上腺有囊肿，已经8年了。\\
\textcolor{my_blue}{医生}：您好！您最近有没有出现任何不适，比如体重变化、乏力或是血压波动等症状？\\
\textcolor{my_yellow}{患者}：没有特别感觉，只是例行检查时发现的。\\
\textcolor{my_blue}{医生}：您之前有做过肾上腺超声检查吗？\\
\textcolor{my_yellow}{患者}：没有，我没有做过肾上腺超声。\\
\textcolor{my_blue}{医生}：那您有做过上腹部增强CT或MRI检查吗？\\
\textcolor{my_yellow}{患者}：根据上个月CTA检查，我双肾有多发的类圆形囊性密影，右肾的较大者约1cm，右肾上腺有一个直径3cm的类圆形肿块。\\
\textcolor{my_blue}{医生}：好的，那您有没有进行过血肾上腺激素检查或尿肾上腺激素检查？\\
\textcolor{my_yellow}{患者}：没有，我没有进行过这些激素检查。\\
\textcolor{my_blue}{医生}：明白了。请问您有没有高血压或其他既往病史？\\
\textcolor{my_yellow}{患者}：我有高血压的病史，但没有糖尿病、冠心病、脑血管疾病等慢性病史。\\
\textcolor{my_blue}{医生}：了解了，您的高血压是怎么控制的？是否有在用药物治疗？\\
\textcolor{my_yellow}{患者}：我一直在用药控制，血压比较稳定。\\
\textcolor{my_blue}{医生}：根据您目前提供的信息，结合您的高血压史和CTA检查结果，我们不能排除肾上腺肿瘤的可能性。虽然目前没有明确的肾上腺激素分泌异常的症状，但为了确诊，我们需要进行血肾上腺激素检查和尿肾上腺激素检查，以及可能的上腹部增强MRI检查。这些检查将有助于我们确定肿瘤的性质和功能活性。
\end{CJK*}

\end{multicols}
\end{tcolorbox}
}
\caption{
Case of dataset \ours
}
\label{fig:datasetcase}
\end{figure*}

\begin{figure*}[!htp]
\centering
\begin{tcolorbox}[center,breakable,title=Template for Converting Question-Answer Pairs from RJUA-QA into Multi-turn Dialogues ]

\begin{center}
\Large \textbf{English}
\end{center}

Your task is to adapt the provided single-turn medical consultation dialogue into a realistic multi-turn conversation between a physician and a patient. The multi-turn dialogue should be entirely based on the single-turn dialogue without adding any new information. It should start with the patient presenting a question, followed by the physician's inquiries, and end with a specific diagnosis from the physician. No treatment suggestions or follow-up actions are necessary.\\

Adhere to the following guidelines when creating the dialogue:\\
1. The dialogue should start with the patient describing the initial symptoms, followed by the physician asking questions to obtain more detailed information.\\
2. Each question from the physician and response from the patient should be closely related and based solely on the information already provided by the patient.\\
3. Avoid introducing new symptoms, tests, or diseases in the dialogue, strictly adhering to the symptoms initially described by the patient.\\
4. The ultimate goal is to establish a specific diagnosis, with the conversation ending when the physician delivers the definitive diagnosis. No treatment suggestions are required.\\

Here is an example of a single-turn medical consultation dialogue:\\
Patient: \texttt{\{\{ QUESTION $q$ \}\}}\\
Physician: \texttt{\{\{ DISEASE $d$ \}\}}\\

Please create a realistic multi-turn physician-patient dialogue based on the provided information, ending with a clear diagnosis from the physician.\\

\begin{center}
\Large \textbf{Chinese}
\end{center}

\begin{CJK*}{UTF8}{gbsn} 
您的工作是利用提供的单轮医疗咨询对话内容，改编成一个符合实际的多轮医生与患者间的对话。多轮对话应当完全基于单轮对话中的信息，不添加任何新信息。多轮对话开始于患者提出问题，继续通过医生的逐个问题来探究，并最终以医生对患者的具体诊断病症结束，无需提供治疗建议或后续行动。\\

遵守以下要点生成对话：\\
1. 开始时患者描述初始症状，之后医生通过询问得到更多详细信息。\\
2. 每个医生的提问和患者的答复都紧密关联，仅基于患者已提供的信息。\\
3. 避免在对话中引入新的症状、检查或疾病，且应当严格遵循患者原有的症状描述。\\
4. 最终目标是确立一个具体的诊断，并以医生的决定性诊断作为对话结束。不需要给出治疗建议。\\

以下是单轮医疗咨询对话示例：\\
患者：\texttt{\{\{ QUESTION $q$ \}\}}\\
医生：\texttt{\{\{ DISEASE $d$ \}\}}\\

请根据提供的信息创建一个逼真的多轮医患对话，并以医生明确的诊断结束。
\end{CJK*}

\end{tcolorbox}
\caption{
This template is needed to transform simple medical question and answer pairs into interactive, multi-turn dialogues that more accurately reflect real-life physician-patient interactions. It is represented as \(M = \textit{G}_M(q , d)\), where $\textit{G}$ denotes the GPT-4 turbo API.
}
\label{fig:ruleprompt}
\end{figure*}

\begin{figure*}[!htp]
\centering
\begin{tcolorbox}[center,breakable,title=Template for Transforming Diagnosis Dialogues into Rule-Based Dialogues]

\begin{center}
\Large \textbf{English}
\end{center}
Adjust the conversation between the physician and the patient according to the following rules. The goal is to establish a specific diagnosis. Start with the patient's description of their condition and end with the physician's definitive diagnosis. No treatment recommendations are needed. The results of any completed tests must be provided by the patient, and the physician can only obtain relevant information through previous dialogue and proceed with further inquiries.\\

When the physician is inquiring, the following rule must be followed:\\
\texttt{\{\{ RULE\_PHYSICIAN $M_\textit{i}$\}\}}\\
When the patient answers the physician's questions, the following rules must be followed:\\
1. Answer the physician's questions completely and honestly. If there are no related symptoms or test results, honestly state none. Do not fabricate or supplement with unperformed test results.\\
2. The patient's goal is to confirm or rule out a disease. Do not perform additional examination actions, end the conversation after obtaining a diagnosis, and if lacking diagnostic evidence, end the conversation without continuing to obtain evidence.\\
3. If there is a clear history of medication, surgery, past illnesses, or marriage and childbirth, it should be retained in the conversation.\\

Original physician-patient dialogue:\\
\texttt{\{\{ DIALOGUES $r_\textit{i}$ \}\}}\\

Adjusted physician-patient dialogue:
\vspace{1em} 

\begin{center}
\Large \textbf{Chinese}
\end{center}
\begin{CJK*}{UTF8}{gbsn} 
将医生和患者对话按照下列规则调整，最终目标是确立具体的诊断。以患者描述自身情况开始，以医生的决定性诊断作为对话结束，不需要给出任何治疗建议。已做的检查结果需由患者提出，医生只能根据前面的对话获得相关信息并进行下一步询问。\\

医生在进行询问时，需遵循规则：\\
\texttt{\{\{ RULE\_PHYSICIAN $M_\textit{i}$ \}\}}\\
患者在回答医生提问时，需遵循规则：\\
1.完整且诚实回答医生提问。若无相关症状或检查结果诚实说明没有，禁止捏造，不得补充没有进行的检查结果。\\
2.患者目标是，确诊或排除疾病。不得进行额外的检查行为，获得诊断后结束对话，若缺少确诊证据也结束对话，不能在获得证据后继续对话。\\
3.有明确的用药史、手术史、既往史、婚育史相关需保留对话。\\

原始医生患者对话:\\
\texttt{\{\{ DIALOGUES $r_\textit{i}$ \}\}}\\

调整后的医生患者对话:
\end{CJK*}

\end{tcolorbox}
\caption{This template aims to guide the transformation of multi-round dialogue $M$ into rule-based dialogue $M^r$. It is represented as \(M^r = \textit{G}_p(M_\textit{i}, r_\textit{i})\), where $\textit{G}$ denotes the GPT-4 turbo API.}
\label{fig:ruleprompt2}
\end{figure*}

\end{document}